\documentclass[10pt,twocolumn,letterpaper]{article}

\usepackage{cvpr} 

\usepackage{graphicx}
\usepackage{amsmath}
\usepackage{amssymb}
\usepackage{booktabs}
\usepackage{multirow}
\usepackage{enumitem}
\usepackage{bbding}
\setlist{nolistsep}
\usepackage{xcolor}

\usepackage[pagebackref,breaklinks,colorlinks]{hyperref}

\newcommand\blfootnote[1]{%
	\begingroup
	\renewcommand\thefootnote{}\footnote{#1}%
	\addtocounter{footnote}{-1}%
	\endgroup
}

\usepackage[capitalize]{cleveref}
\crefname{section}{Sec.}{Secs.}
\Crefname{section}{Section}{Sections}
\Crefname{table}{Table}{Tables}
\crefname{table}{Tab.}{Tabs.}

\def\cvprPaperID{3860} 
\def\confName{CVPR}
\def\confYear{2023}

\begin{document}

\title{KiUT: Knowledge-injected U-Transformer for Radiology Report Generation}

\author{Zhongzhen Huang\textsuperscript{1,2}, \quad Xiaofan Zhang\textsuperscript{1,2} \Envelope , \quad Shaoting Zhang \textsuperscript{2,3} \\
\textsuperscript{1}Shanghai Jiao Tong University \quad
\textsuperscript{2}Shanghai AI Laboratory \quad \textsuperscript{3}SenseTime Research \quad\\
\tt\small{\{huangzhongzhen, xiaofan.zhang\}@sjtu.edu.cn, zhangshaoting@pjlab.org.cn}\\
}

\maketitle
\begin{abstract}
Radiology report generation aims to automatically generate a clinically accurate and coherent paragraph from the X-ray image, which could relieve radiologists from the heavy burden of report writing. Although various image caption methods have shown remarkable performance in the natural image field, generating accurate reports for medical images requires knowledge of multiple modalities, including vision, language, and medical terminology. We propose a Knowledge-injected U-Transformer (KiUT) to learn multi-level visual representation and adaptively distill the information with contextual and clinical knowledge for word prediction.
In detail, a U-connection schema between the encoder and decoder is designed to model interactions between different modalities.
And a symptom graph and an injected knowledge distiller are developed to assist the report generation. 
Experimentally, we outperform state-of-the-art methods on two widely used benchmark datasets: IU-Xray and MIMIC-CXR. Further experimental results prove the advantages of our architecture and the complementary benefits of the injected knowledge.
\end{abstract}
\blfootnote{\Envelope \ Corresponding Author. }
\section{Introduction}
\label{sec:intro}
\begin{figure}[t]
\centering
\includegraphics[width=0.9\linewidth]{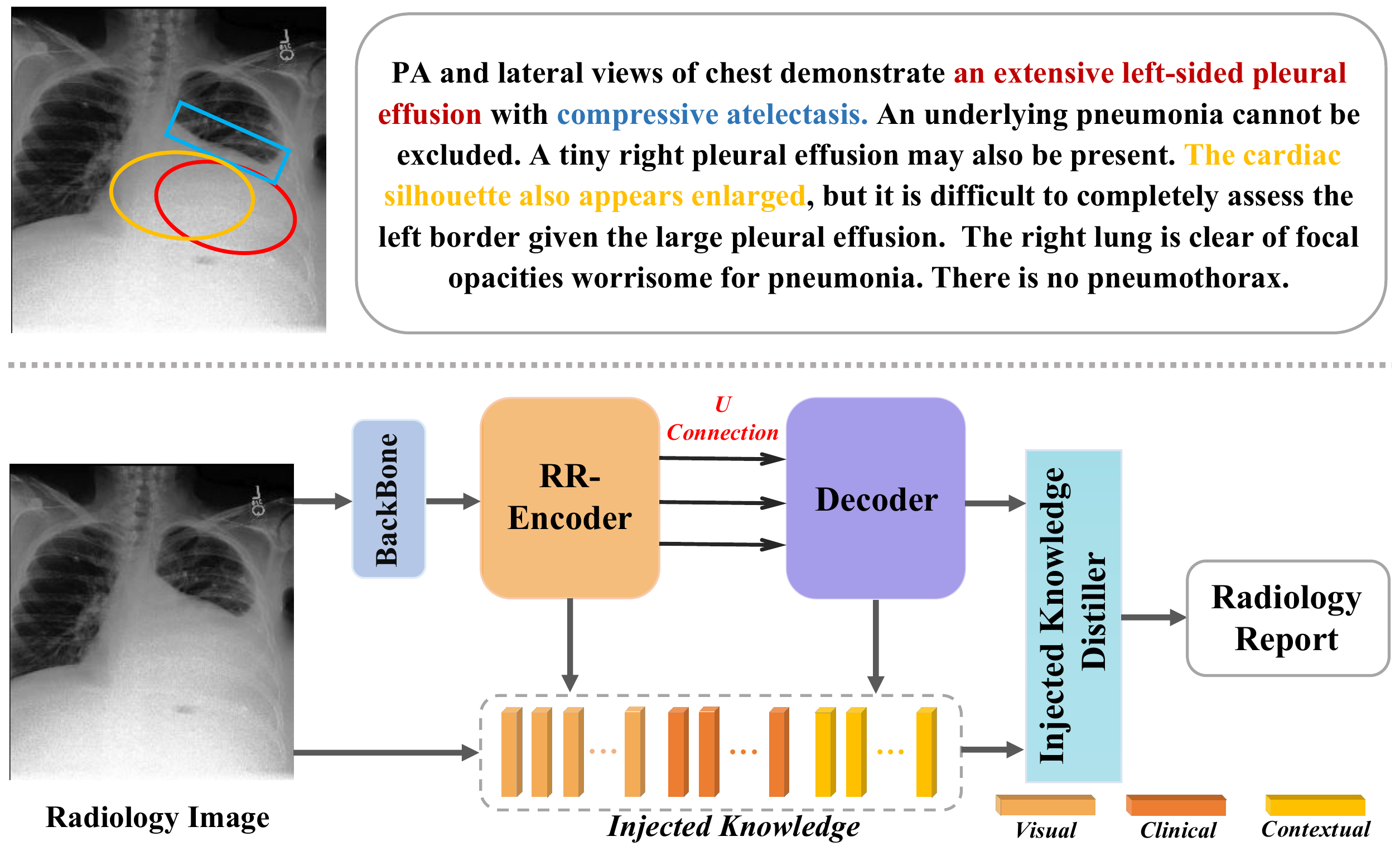}
\caption{
A transformer architecture with U-connection is adopted to generate reports from radiology images. The process involves injecting and distilling visual, clinical, and contextual knowledge
The color labels in the image and report represent the different abnormal regions and their corresponding description, respectively.}

\vspace{-1em}
\label{fig:introduction} 
\end{figure}
Radiology images (\eg, chest X-ray) play an indispensable role in routine diagnosis and treatment, and the radiology reports of images are essential in facilitating later treatments. Getting a hand-crafted report is a time-consuming and error-prone process. Given a radiology image, only experienced radiologists can accurately interpret the image and write down corresponding findings. Therefore, automatically generating high-quality radiology reports is urgently needed to help radiologists eliminate the overwhelming volume of radiology images.
In recent years, radiology report generation has attracted much attention in the deep learning and medical domain. The encoder-decoder architecture inspired by neural machine translation\cite{sutskever2014sequence} has been widely adopted by most existing methods\cite{jing2017automatic,xue2018multimodal,yuan2019automatic,liang2017recurrent}. With the recent advent of the attention mechanism, the architecture's capability is greatly ameliorated. 

Despite the remarkable performance, these models restrained themselves in the methodology of image caption\cite{zhang2021rstnet,cornia2020meshed,cornia2019show,xu2015show,vinyals2015show}, and suffer from such data biases:
1) the normal cases dominate the dataset over the abnormal cases; 2) the descriptions of normal regions dominate the entire report. 
Recently, some methods have been proposed to either alleviate case-level bias by utilizing posterior and prior knowledge \cite{liu2021exploring} 
or relieve the region-level bias by distilling the contrastive information of abnormal regions \cite{liu2021contrastive}.

Thus, in the medical field's cross-modal task, a model needs to not only capture visual details of every abnormal region but also consider the interaction between the visual and textual modalities among different levels.
Moreover, external clinical knowledge is required to achieve the radiologist-like ability in radiology image understanding and report writing. 
The external knowledge, \eg, the clinical entities and relationships, could be pre-defined by experts or mined from medical documents. However, directly adopting the knowledge brings inconsistencies due to the heterogeneous context embedding space\cite{li2022cross}.
And too complex knowledge may be prone to distract the visual encoder and divert the representation\cite{liu2021exploring}.

Instead of using external knowledge to augment the feature extraction like previous approaches\cite{liu2021exploring,zhang2020radiology}, we propose to introduce the injected knowledge in the final decoding stage.
A graph with the clinical entities, \ie, symptoms and their relationships, is constructed under the guidance of professional doctors. 
These entities have homogeneous embedding space with the training corpus, and this signal could be injected smoothly with visual and contextual information. 
We further design the Injected Knowledge Distiller on top of the decoder to distill contributive knowledge from visual, contextual, and clinical knowledge. 



Following these premises, we explore a novel framework dubbed as \textit{\textbf{K}nowledge-\textbf{i}njected  and \textbf{U}-\textbf{T}ransformer} (KiUT) to achieve the radiologist-like ability to understand the radiology images and write reports. As \cref{fig:introduction} shows, it consists of a Region Relationship Encoder and Decoder with U-connection architecture and Injected Knowledge Distiller.

Our contributions can be summarized as follows:
\begin{itemize}[noitemsep, topsep=0pt]
\item We propose a novel model following the encoder-decoder architecture with U-connection that fully exploits different levels of visual information instead of only one single input from the visual modality. In our experiments, the U-connection schema presents improvement not only in radiology report generation but also in the natural image captioning task. 

\item Our proposed model injects clinical knowledge by constructing a symptom graph, combining it with the visual and contextual information, and distilling them when generating the final words in the decoding stage.

\item The Region Relationship Encoder is developed to restore the extrinsic and intrinsic relationships among image regions for extracting abnormal region features, which are crucial in the medical domain.  

\item We evaluate our approach on two public radiology report generation datasets, IU-Xray\cite{demner2016preparing} and MIMIC-CXR\cite{johnson2019mimic}. \textit{KiUT} achieves state-of-the-art performance on the two benchmark datasets.
\end{itemize}

\begin{figure*}[h]
\centering
\scalebox{1}{
\includegraphics[width=\linewidth]{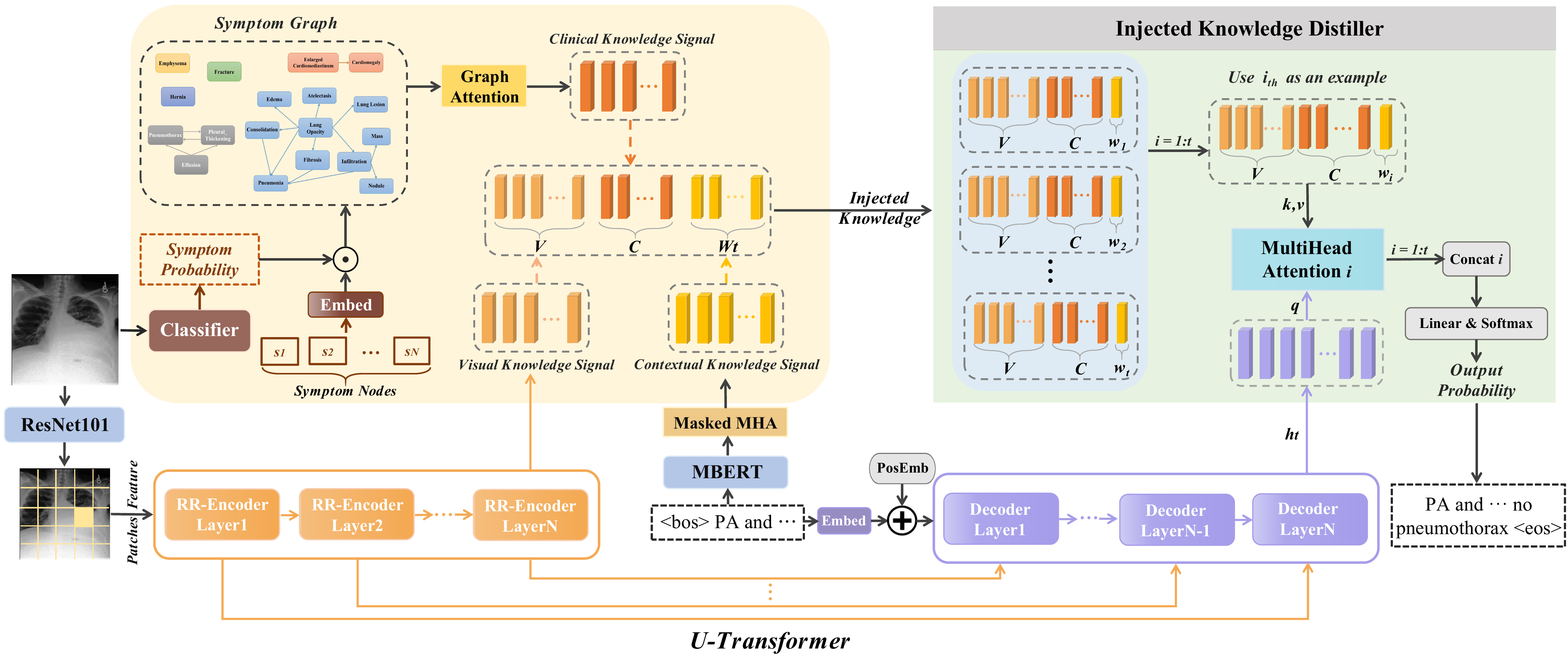}
}
\caption{The overall architecture of \textit{KiUT}: 1) U-Transformer with U-connection between Visual RR-Encoder and Decoder, 2) Injected Knowledge Distiller that handles three types of injected knowledge. There are seriatim elaborate descriptions of these modules in Section~\ref{Method}. 
}
\vspace{-4pt}
\label{fig:arch}
\end{figure*}
\section{Related Work}
\subsection{Image Caption}
Generating radiology reports is similar to the image caption task, which aims to describe the content of a given natural image.
Based on the encoder-decoder architecture, the development of image caption models\cite{xu2015show,vinyals2016show,ji2021improving,rennie2017self,lu2017knowing} has advanced in leaps and bounds.  

With the good performance of Transformer\cite{vaswani2017attention} in both computer vision and natural language processing field, a broad collection of transformer based methods have been explored to ameliorate the performance of Image Caption.
A combined bottom-up and top-down attention mechanism was introduced by \cite{anderson2018bottom} to calculate the relationships between objects and other salient image regions when extracting visual representation. 
\cite{cornia2020meshed} proposed a meshed-memory transformer that represents visual features incorporating the relationships between image regions.
In the work\cite{zhang2021rstnet}, Zhang~\etal enhanced visual representations in the attention module with relative geometry features and used adaptive attention to predict visual and non-visual words.
\textit{GRIT} \cite{nguyen2022grit} adopts a transformer architecture to fuse the objects' feature and contextual features for better captioning. 
Although these methods are effective, they still need to take full advantage of the visual information and consider the interaction between the encoder and decoder. To explore the relationship between these two modalities, \cite{cornia2020meshed} tried complex mesh-like connectivity at the decoding stage. 
In this paper, we try a simple and reasonable schema to establish the connection between the encoder and decoder.

\subsection{Radiology Report Generation}
Radiology reports contain multiple descriptive sentences about radiology images. 
As the transfer and extension of Image Caption in the medical field, radiology report generation not only puts forward higher requirements on the length of generated reports but also presents greater challenges on the accuracy of long contextual descriptions. 
Much work has made some advancements in this task. \cite{li2018hybrid} employed a hierarchical decision making procedure and generated normal and abnormal sentences by multi-agent systems, respectively.
Liu~\etal \cite{liu2019clinically}  presented a hierarchical generation framework which first predicts the related topics of the input, then generates sentences according to the topics. For the latter studies, transformer\cite{vaswani2017attention} has been successfully applied as the model design rationale. \cite{hou2021ratchet} developed a transformer-based architecture and used attention to localize important regions in the image for generation. Chen~\etal\cite{chen2022cross} improved the performance by incorporating the transformer based model with memory matrix which is designed to align cross modal features. 

Recently, some work \cite{li2022auxiliary,li2019knowledge,li2022cross,wang2022cross,zhang2020radiology,liu2021exploring} began to explore assisting report generation with additional knowledge. Wang~\etal\cite{wang2022cross} utilized a shared cross-modal prototype matrix as external knowledge to record the cross-modal prototypes and embed the cross-modal information for the reports generation. \cite{zhang2020radiology} designed a pre-constructed medical graph based on prior knowledge from chest findings, which allows dedicated feature learning for each disease finding. 
The graph improves the models’ capability to understand medical domain knowledge. Inspired by the idea, Liu~\etal explored and distilled posterior and prior knowledge in the work\cite{liu2021exploring}, where the posterior knowledge is visual information and the prior knowledge is the medical graph and retrieval reports. Instead of applying knowledge for visual feature extraction, we distill knowledge in the decoding stage. The report generated by our model is based on the distilled knowledge from multiple modalities.


\vspace{-4pt}
\section{Methodology}
\label{Method}
In the radiology report generation task, given a 2D radiology image $I$, the model is required to interpret the image and generate a descriptive radiology report $R=\lbrace{y_1,y_2,\dots,y_{N_R}\rbrace}$ , where $y_i$ is the word token of the report and $N_R$ is the length of the report. 
The entire recursive generation process can be formulated as follows:
\begin{equation}
p(R \mid I)=\prod_{t=1} p\left(y_{t+1} \mid y_1, \ldots, y_{t}, I\right)
\end{equation}
And the model is generally optimized by minimizing the cross-entropy loss:
\begin{equation}
L_{\mathrm{CE}}(\theta)=-\sum_{i=1}^{N} \log \left(p_{\theta}\left(\mathbf{y}_{n}^{*} \mid \mathbf{y}_{1: n-1}^{*}\right)\right)
\end{equation}
where $R^*=\lbrace{y^*_1,y^*_2,\dots,y^*_{N_R}\rbrace}$ is the ground truth report.

In this section, we will introduce the framework of the proposed \textit{KiUT}. \cref{fig:arch} shows the overall architecture of our proposed \textit{KiUT}. Our model can be conceptually divided into three core components: Cross-modal U-Transformer, Injected Knowledge Distiller, and Region Relationship Encoder. We will introduce their details in turn.

\subsection{Cross-modal U-Transformer}
\begin{figure}[h]
\centering
\scalebox{0.7}{
\includegraphics[width=\linewidth]{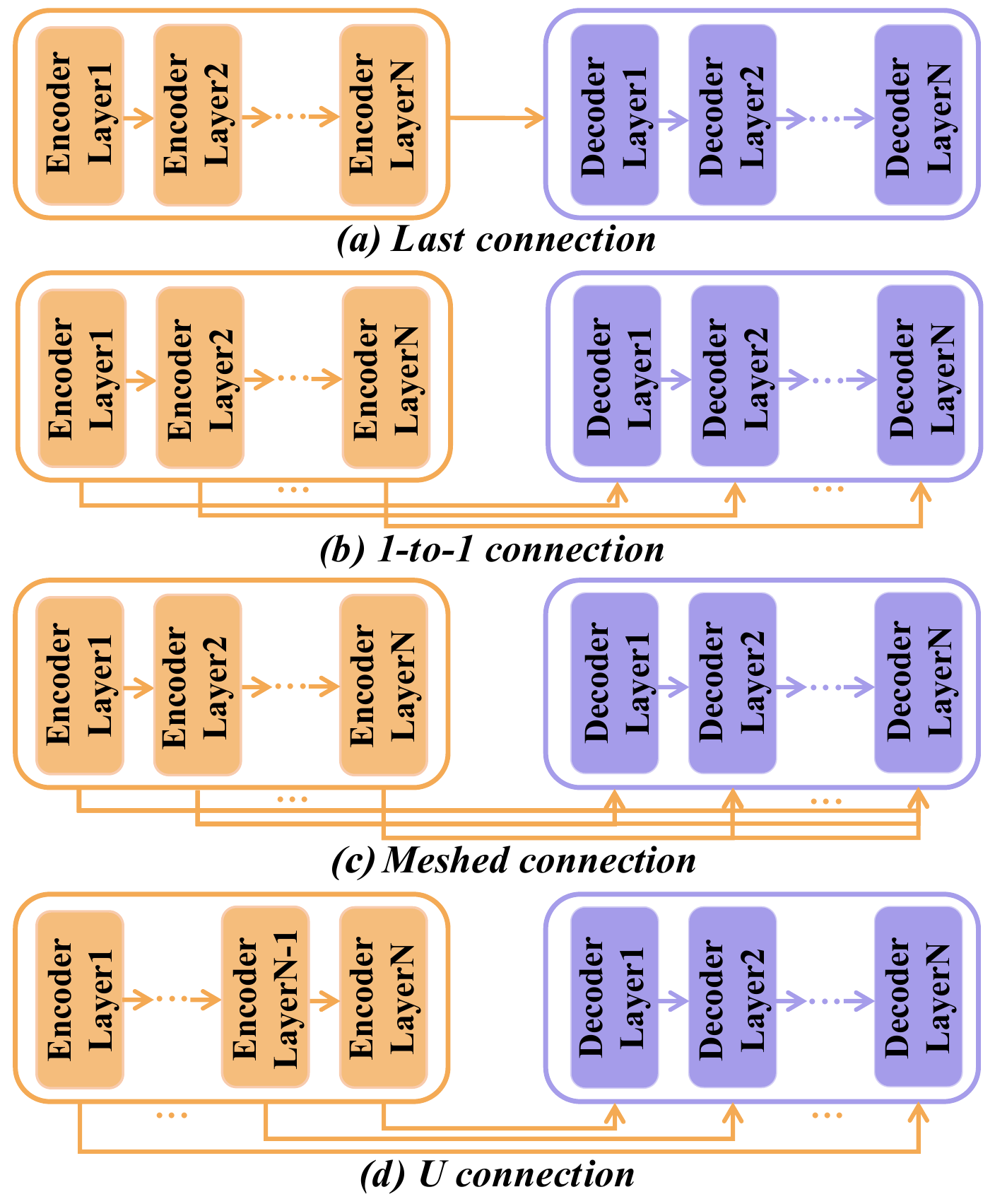}
}
\caption{Connection schemas between the encoder and decoder.}
\label{fig:U_1_last}
\end{figure}

For cross-modal tasks, it is particularly important to obtain proper interaction between different modalities. As shown in \cref{fig:U_1_last}, compared with the previous work using only the last output of the encoder, or proposing a meshed connection\cite{cornia2020meshed}, we design a U-transformer architecture with much fewer parameters than the meshed connection. Specifically, we construct a U-connection that connects the layers of the encoder and decoder.

The visual encoder captures details in X-ray images by the multi-head self-attention that naturally aggregates task-relevant features. 
The decoder aggregates the visual features throughout all layers via U-connection.
And the output features of the last decoder layer are sent to the Injected Knowledge Distiller for generating the words in the report.


Assume that the number of layers of the proposed encoder and decoder is $N$. The output of each layer of the encoder can be formulated as $\lbrace{\tilde{x}_1,\dots,\tilde{x}_i,\dots,\tilde{x}_N}\rbrace$, where $\tilde{x}_i$ is defined as the $i$-th layer's output of our Visual RR-Encoder (\cref{sec:VRRE}).
In our proposed architecture, the output of the $i$-th encoder layer will be input to the $(N-i+1)$-th decoder layer. The specific process can be formulated as: 
\begin{gather}
    \lbrace{\tilde{x}_1,\dots,\tilde{x}_N}\rbrace = \operatorname{RR-Encoder}(I)\\
    \hat{x}_i=\tilde{x}_{N-i+1}   
\end{gather}
where $\hat{x}_i$ is the input of the $i$-th decoder layer.

After receiving the visual features $\lbrace{\hat{x}_1,\dots,\hat{x}_N}\rbrace$ from the encoder, the decoder generates the hidden state $h_t$ to predict the current word $y_{t+1}$ through the word sequence features $\hat{w}_{t}$ in the decoding step t+1. Given the generated words $y_{1:t} = \lbrace{y_1, \dots, y_t}\rbrace$, the word sequence features are obtained by words embeddings $w_t$ combining with its position encodings $e_t$. The process of the decoder is as follows:
\begin{gather}
e_{t} = \operatorname{PosEncoding}(y_{1:t})\\
\hat{w}_{t} = w_{t}+e_{t} \\
h^i_t= \begin{cases}\operatorname{DecoderLayer_i}(\hat{w}_{t},\hat{x}_1) & i=1 \\ \operatorname{DecoderLayer_i}(h^{i-1}_t,\hat{x}_i) & i>1\end{cases}
\end{gather}
where the $\operatorname{DecoderLayer_i}$ and $h^i_t$ represent the $i$-th DecoderLayer and its output, the final hidden state $h_t = h^N_t$. $\operatorname{DecoderLayer}$ and $\operatorname{PosEncoding}$ follows the operations of \emph{Decoder} and \emph{Position encoding} in \cite{vaswani2017attention}.

\subsection{Injected Knowledge Distiller}
Radiologists draw on different aspects of medical knowledge when writing reports. To make the process of report generation conformance to the real scenario and generate more accurate reports, our model incorporates injected knowledge signals from three aspects, \ie, visual knowledge, contextual knowledge, and clinical knowledge.


\textbf{Visual knowledge signal} contains the information of the radiology image $I$. $\tilde{X}$ is the output of the encoder's last layer, namely $\tilde{X}=\tilde{x}_N$.

\label{contextual}
\textbf{Contextual knowledge signal} is the information of the generated words extracted from \emph{MBert} with a masked attention module. Specifically, to endow \emph{MBert} with partial medical language knowledge, we adopt a pre-trained \emph{Bert} model\cite{devlin2018bert} and finetune it on the reports in train split of the specific dataset\cite{johnson2019mimic,demner2016preparing}. And a $\operatorname{MaskedMHA}$\cite{vaswani2017attention} module is applied after \emph{MBert} to obtain linguistic features of the generated report words. The process of extracting contextual knowledge signal can be defined as:
\begin{gather}
\tilde{W}_{t}=\operatorname{MaskedMHA}(MBert(y_{1:t})+e_{t})\end{gather}
where $\tilde{W}_{t}\in\mathbb{R}^{t\times d}$ and $y_{1:t}$ is the generated words sequence.

\begin{figure}[h]
\centering
\scalebox{1}{
\includegraphics[width=0.9 \linewidth]{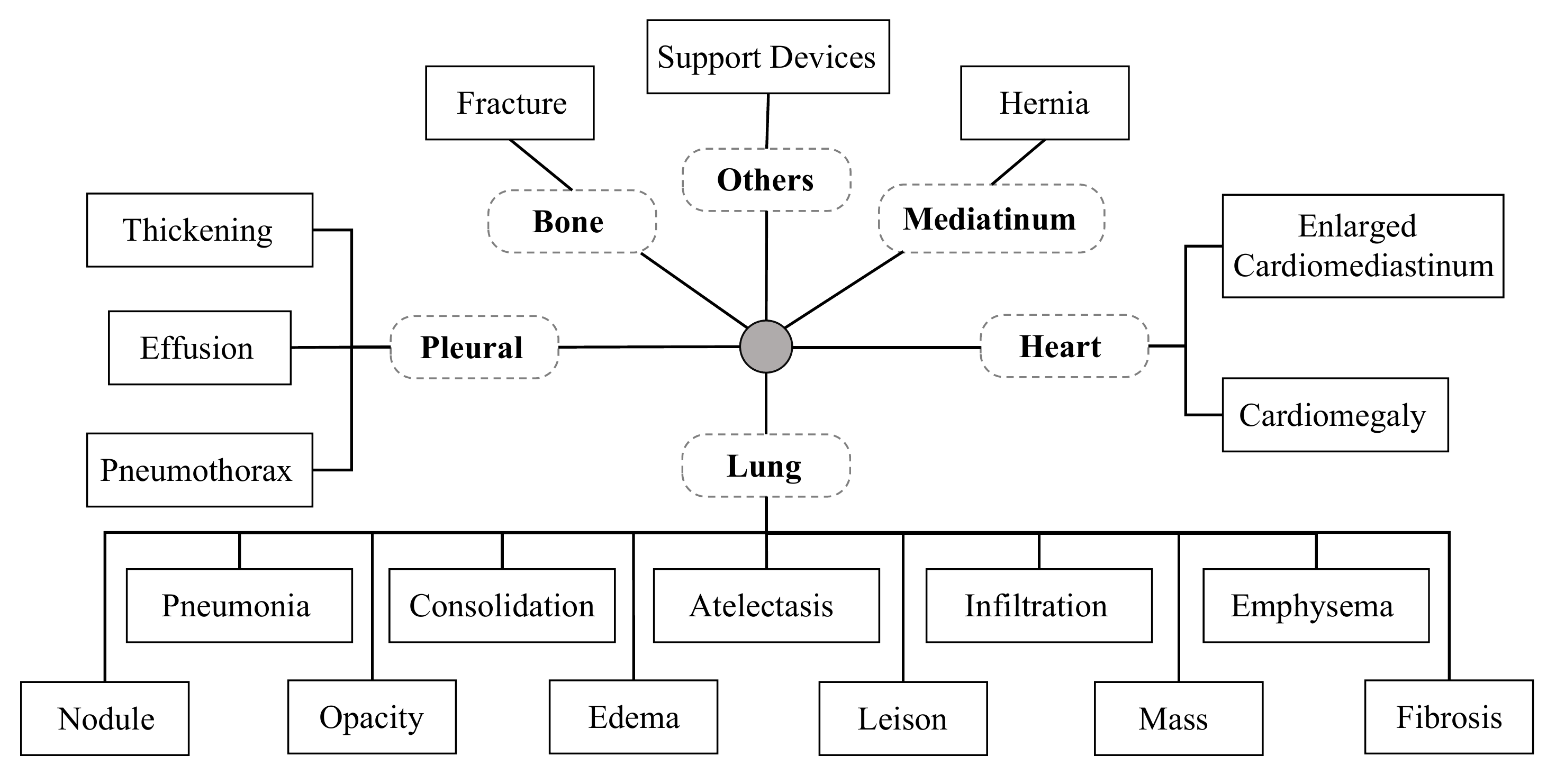}
}
\caption{The symptoms graph is based on the correlation, characteristics and occurrence location of symptoms.}
\label{fig:symptoms}
\vspace{-0.5em}
\end{figure}

\textbf{Clinical knowledge signal} is essential to the radiology report generation task. 
Imaging exam is scheduled for a specific purpose in the medical field. For example, the chest X-ray is a part of the physical exam that could show the size, shape, and location of the heart, lungs, bronchi, pulmonary arteries, \etc.
Therefore, we could design a clinical symptom graph to inject real-world medical knowledge into the model. The graph shown in \cref{fig:symptoms} is developed according to the professional perspective of radiology images, taking into account symptoms correlation, symptom characteristics, occurrence location, \etc. The feature of each node $sf_i \in \mathbb{R}^d$ in the graph is the word embedding of each symptom derived from the \emph{MBert}.

For every radiology image $I$, the probability distribution $sp_i$ of each symptom in the graph is calculated with the pre-trained \emph{classification model}\cite{Cohen2022xrv}.
Then, the symptom graph $G$ is initialized according to the distribution, and the graph attention mechanism (GAT) is utilized to calculate the clinical knowledge signal $\tilde{C}$. The process is as follows:
\begin{gather}
sg_i = sp_i \odot sf_i , g_i \in G\\ \tilde{C}=\operatorname{GAT}(G)\end{gather}
where $g_i$  represents $i$th node of the symptom graph $G$.

The clinical knowledge signal indicates the possible findings that may need to be written in the report. And these findings reflect the radiologists' first impression of the image based on their professional knowledge.

\textbf{Knowledge Distiller}
To make our model achieve the radiologist-like ability to generate reports and align with the real  scenario, we propose an Injected Knowledge Distiller to distill useful information from the above knowledge signals. The process is formally defined as follows:
\begin{gather}
\tilde{F}_i = \left[\tilde{X}, \tilde{C}, \tilde{w}_{i}\right] \ \ i \in [1,t], \ \tilde{w}_{i} \in \tilde{W}_t
\\\operatorname{MHA}_i(h_{t},\tilde{F}_i) =\operatorname{Softmax}\left(\frac{QK_i^{T}}{\sqrt{d_n}}\right)V_i
\\Q = h_{t} \mathrm{W}^{Q}, K_{i}=\tilde{F}_i\mathrm{W}^{K}, V_{i}=\tilde{F}_i \mathrm{W}^{V}
\\\tilde{h}_{t} = \operatorname{Concat}(\operatorname{MHA}_1(h_{t},\tilde{F}_1), \cdots, \operatorname{MHA}_t(h_{t},\tilde{F}_t))
\\{y}_{t+1} \sim p_{t+1} = \operatorname{Softmax}\left(\tilde{h}_{t} \mathrm{W}_{p}+\mathrm{b}_{p}\right)
\end{gather}
where $\tilde{w}_{i}\in \tilde{W}_{t}$, $\mathrm{W}^{Q},\mathrm{W}^{K}, \mathrm{W}^{V}, \mathrm{W}_{p}$ and $\mathrm{b}_{p}$ are learnable parameters. ${y}_{t+1} \sim p_{t+1}$ is the probability of the generated word at step $t+1$.

\subsection{Region Relationship Encoder}
\label{sec:VRRE}
To get the visual representation, the first step is to extract features from the radiology image. Following \cite{chen2020generating,wang2022cross,chen2022cross}, we adopt pre-trained convolutional neural networks to extract the basic visual feature $X$ from a radiology image $I$. Generally, the image is decomposed into $S$ patches of equal size by flattening the extracted feature by row. In brief, $X = \{x_i\}_{i=1}^{S}$ is the source input for the subsequent modules, and $x_i \in \mathbb{R}^d$.

After getting a sequence of the visual features $X$ extracted from radiology image regions, it is crucial to explore the extrinsic and intrinsic relationships among these regions for better visual representations, which are important for understanding radiology images. For example, the extrinsic relationship: \textit{the heart is located below the left lung} and the intrinsic relationship: \textit{enlarged cardiomediastinum is usually associated with some left lung imaging manifestations}.
Although \emph{self-attention}\cite{vaswani2017attention} can be used to model the relationship among image regions, it can only calculate the similarity between image region features\cite{cornia2020meshed}. Therefore,
we propose a region relationship encoder to explore the extrinsic and intrinsic relationships among the regions.

For the flattened image region features, the loss of the extrinsic relationships is inevitable, \eg, the spatial information. Therefore, we incorporate relative geometry information into \emph{self-attention} to take into account the extrinsic relationships among regions. 
We first calculate the center coordinate $(x_i,y_i)$, width $w_i$, and height $h_i$ of a region $i$, the specific calculation as follows: 
\begin{gather}
\left(x_{i}, y_{i}\right) =\left(\frac{x_{i}^{\min }+x_{i}^{\max }}{2},\frac{y_{i}^{\min }+y_{i}^{\max }}{2}\right)\\w_{i} =\left(x_{i}^{\max }-x_{i}^{\min }\right)+1\\ h_{i} =\left(y_{i}^{\max }-y_{i}^{\min }\right)+1
\end{gather}
where $(x_i^{max},y_i^{max})$ and $(x_i^{min},y_i^{min})$ are the position coordinate of the upper left corner and the lower right corner of the grid $i$, respectively.
Following the computation of region relative geometry features in \cite{herdade2019image,guo2020normalized},
the relative geometry relationship of regions can be calculated by:
\begin{equation}
\begin{aligned}
rg_{i j}=\Big(
\log \left(\frac{\left|x_{i}-x_{j}\right|}{w_{i}}\right)&,
\log \left(\frac{\left|y_{i}-y_{j}\right|}{h_{i}}\right), \\
\log \left(\frac{w_{i}}{w_{j}}\right)&,
\log \left(\frac{h_{i}}{h_{j}}\right)\Big)^T
\end{aligned}
\end{equation}
where $rg \in \mathbb{R}^{S \times S \times 4}$.
And the extrinsic relationship $ER_{ij}$ between region $i$ and $j$ can be obtained by:
\begin{equation}
ER_{i j}=\operatorname{ReLU}(\mathrm{w_{g}}^T FC(rg_{ij}))
\end{equation}
where $\mathrm{w_g}$ is a learnable parameter, $FC$ is  a fully-connected layer and $\operatorname{ReLU}$ is an activation for zero trimming. $ER \in \mathbb{R}^{S \times S}$ is the representation of the extrinsic relationship.

To represent the intrinsic relationship among regions, we use additional learnable matrices and extend the set of keys and values in \emph{self-attention} with them\cite{cornia2020meshed}. This operation is defined as:
\begin{gather}
Keys,Values = \mathrm{W}_{k} X, \mathrm{W}_{v} X\\
K_{I}=\left[Keys,\boldsymbol{M}_{k}\right] ,V_{I}=\left[Values, \boldsymbol{M}_{v}\right] 
\end{gather}
where $\boldsymbol{M}_{k}$ and $\boldsymbol{M}_{v}$ are learnable matrices with $M_N$ rows, $X$ is the input to \emph{self-attention}, $\mathrm{W}_{k}$ and $\mathrm{W}_{v}$ are are learnable parameters. $[\cdot,\cdot]$ stands for concatenation operation.

Based on the above, to compensate for the region relationship in the visual encoder, we propose a region relationship augmented self-attention$(\mathcal{RRSA})$ in our encoder. The $\mathcal{RRSA}$ is formally defined as follows:
\begin{gather}\mathcal{RRSA}(Q,K_I,V_I)=\operatorname{Softmax}\left(\frac{QK_{I}^{T}}{\sqrt{d_n}}+ER_I\right)V_{I}
\\
Q = \mathrm{W}_{q} X, ER_I=\left[ER, \boldsymbol{M}_{I}\right]\end{gather}
where $\mathrm{W}_{v}$ and $\boldsymbol{M}_{I}$ are learnable parameters. The visual features $\lbrace{\tilde{x}_1,\dots,\tilde{x}_N}\rbrace$ are obtained by the $\mathcal{RRSA}$ based region relationship encoder.

\section{Experiment}
\subsection{Dataset}
We conduct experiments to evaluate the effectiveness of the proposed \emph{KiUT} on two widely used medical report generation benchmarks, \ie, IU-Xray and MIMIC-CXR. 

\textbf{IU-Xray} \cite{demner2016preparing} from Indiana University is a relatively small but publicly available dataset containing 7,470 chest X-ray images and 3,955 radiology reports. For the majority of reports, there are frontal and lateral radiology images. Following the previous work \cite{chen2020generating,chen2022cross,wang2022cross,liu2021exploring}, we first exclude the samples without \emph{"Findings"} section and follow the widely-used splits to divide the dataset into train (70\%), validation (10\%) and test (20\%) sets.

\textbf{MIMIC-CXR} \cite{johnson2019mimic} provided by the Beth Israel Deaconess Medical Center is a recently released large-scale dataset. The dataset includes 377,110 chest X-ray images and 227,835 reports. For a fair comparison, we adopt the official data splits (\ie, 70\%/10\%/20\% for train/validation/test set). Thus, there are 368,960 in the training set, 2,991 in the validation set, and 5,159 in the test set. 

\begin{figure}[htb]
\centering
\scalebox{1}{
\includegraphics[width=\linewidth]{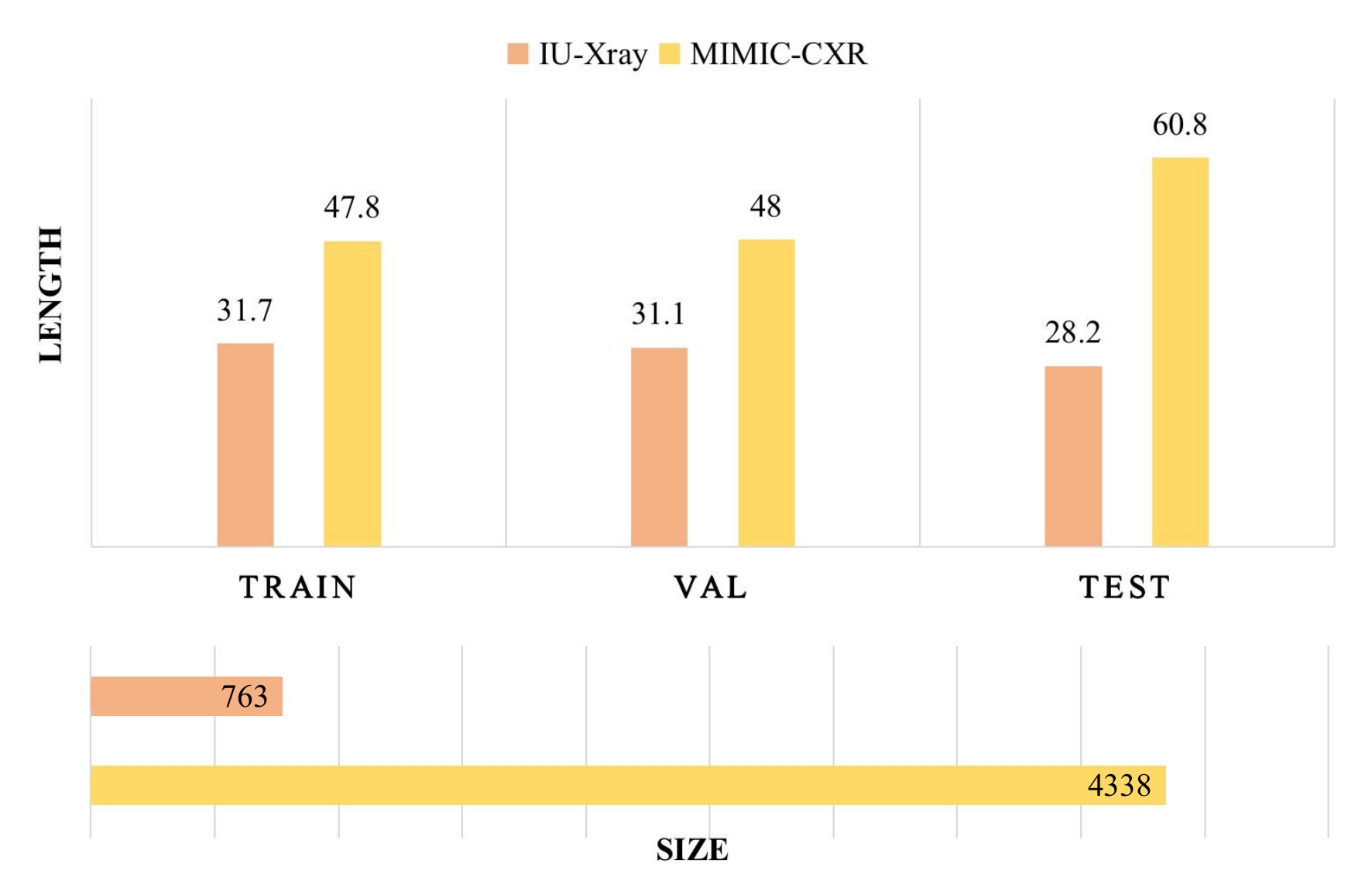}
}
\caption{The size of vocabulary and the average length of reports for two datasets.}
\label{fig:compare}
\end{figure}

\begin{table*}[h!]
  \caption{Performance comparisons of the proposed KiUT with existing methods on the test sets of MIMIC-CXR and IU-Xray datasets with respect to NLG and CE metrics. The best values are highlighted in bold.}
  \label{tab:result_compare}
  \centering
  \begin{tabular}{c|l|c|c|c|c|c|c|c|c|c}
    \hline
    \multicolumn{1}{c|}{\multirow{2}{*}{Dataset}} &
    \multicolumn{1}{c|}{\multirow{2}{*}{Method}} &
    \multicolumn{6}{|c}{NLG Metric} & 
    \multicolumn{3}{|c}{CE Metric} 
    \\\cline {3-11} ~ & ~ & Bleu1 & Bleu2 &Bleu3 &Bleu4 &Meteor & Rouge\_L  & Precision & Recall & F1 \\ \hline \hline
    $ $ &SentSAT+KG\cite{zhang2020radiology} &0.441 &0.291 &0.203 &0.147 &$-$ &0.367 & $-$& $-$ & $-$\\
    $ $  & R2GenCMN\cite{chen2022cross}&0.470&0.304&0.219&0.165&0.187&0.371& $-$ & $-$ & $-$\\
    \textbf{IU-} & PPKED\cite{liu2021exploring}&0.483&0.315&0.224&0.168&0.190& 0.376 & $-$ & $-$ & $-$\\
    \textbf{Xray} & AlignTrans\cite{you2021aligntransformer}& 0.484&  0.313& 0.225& 0.173& 0.204& 0.379  & $-$ & $-$ & $-$\\
  $ $ &Contrastive\cite{liu2021contrastive} &0.492 &0.314 &0.222 &0.169 &0.193 &0.381 & $-$& $-$ & $-$\\
    $ $ & XPRONET\cite{wang2022cross} & 0.525 & 0.357 & \textbf{0.262} &\textbf{0.199}& 0.220 & \textbf{0.411}  &$-$ & $-$ & $-$\\
    \cline {2-11}
  $ $ & Ours& \textbf{0.525} & \textbf{0.360} & 0.251 & 0.185& \textbf{0.242} &0.409 & $-$ & $-$ & $-$\\
      \hline 
      $ $& Up-Down\cite{anderson2018bottom}&  0.317&   0.195&  0.130&  0.092&  0.128 & 0.267 &  0.320 &0.231 &0.238\\
    $ $ & Att2in\cite{rennie2017self}&  0.325&   0.203&  0.136&  0.096 &  0.134 & 0.276  &  0.322 & 0.239 & 0.249\\
    $ $ & R2GenCMN\cite{chen2022cross}&  0.353&    0.218&   0.145&  0.103&   0.142& 0.277  & 0.333 &0.273 &0.276\\
    \textbf{MIMIC} & PPKED\cite{liu2021exploring}& 0.360& 0.224&   0.149&  0.106&  0.149& 0.284 & $-$ & $-$ & $-$\\
    \textbf{CXR} & Contrastive\cite{liu2021contrastive}& 0.350 &0.219 &0.152  &0.109 &0.151 &0.283 &0.352 &0.298 &0.303\\
    $ $ & AlignTrans\cite{you2021aligntransformer}& 0.378&  0.235& 0.156& 0.112& 0.158& 0.283  & $-$ & $-$ & $-$\\
    $ $ & XPRONET\cite{wang2022cross} & 0.344 & 0.215 & 0.146 & 0.105& 0.138 & 0.279  & $-$ & $-$ & $-$\\
    \cline {2-11}
    $ $ & Ours& \textbf{0.393} & \textbf{0.243} & \textbf{0.159} &\textbf{0.113}& \textbf{0.160} &\textbf{0.285} & \textbf{0.371} & \textbf{0.318} & \textbf{0.321}\\
    \bottomrule
  \end{tabular}
\end{table*}

For the reports of these two datasets, we preprocess them by tokenizing and converting all tokens to lower cases and removing the special tokens like digital, non-alphanumeric characters, etc. Finally, we remove the tokens whose frequency of occurrence is less than a specific threshold from the remaining tokens to construct our vocabulary. \cref{fig:compare} shows the statistical visualization of all datasets in terms of the size of vocabulary and the average length of reports.
\subsection{Implementation Details}
Following the previous work\cite{chen2020generating,wang2022cross,chen2022cross}, we adopt the ResNet-101\cite{he2016deep} pertained on the ImageNet\cite{krizhevsky2017imagenet} to extract image region features. The obtained features are further projected into 512-dimension in the shape of $7 \times 7$, \ie $S$ is 49 and $d$ is 512. To ensure consistency with the existing methods setting\cite{wang2022cross,chen2022cross,liu2021exploring}, we utilize paired images of a patient as the input for IU-Xray and one image for MIMIC-CXR. For the encoder-decoder architecture, we follow the implementation of a Transformer-based model and keep the inner structure and parameters untouched. 
To facilitate the method, we use \emph{XRayVison} models, a pre-trained DenseNet-121\cite{huang2017densely} on radiology images, to generate probabilities for the mentioned symptoms graph.

\subsection{Metrics}
To gauge the performance, we employ the widely-used  natural language generation (NLG) metrics and clinical efficacy (CE) metrics. We adopt the standard evaluation protocol to calculate the captioning metrics: BLEU\cite{papineni2002bleu}, METEOR\cite{banerjee2005meteor} and ROUGE-L\cite{lin2004rouge}. And for the clinical efficacy, we apply CheXpert\cite{irvin2019chexpert} to label the generated reports and compare the labeling results with ground truths in 14 categories of diseases through precision, recall and F1.

\begin{table*}[h!]
  \caption{Ablation study of our method on the MIMIC-CXR dataset, which includes the RR-Encoder and IK-Distiller.}
  \label{tab:ablation}
  \centering
  \begin{tabular}{c|c|c|c|c|c|c|c|c|c|c}
    \toprule
    \multirow{2}{*}{Dataset} &
    \multicolumn{2}{|c}{RR-Encoder} & 
    \multicolumn{2}{|c}{IK-Distiller} &
    \multicolumn{6}{|c}{Metric}
    \\\cmidrule(lr){2-3}\cmidrule(lr){4-5}\cmidrule(lr){6-11} ~ & ER & IR & Contextual & Clinical & Bleu1 & Bleu2 &Bleu3 &Bleu4 &Meteor & Rouge\_L  \\ 
    \midrule
    $ $ &  &  & \checkmark & \checkmark &0.373&   0.224&  0.148&  0.106  & 0.137  & 0.278  \\
    $ $ &  & \checkmark & \checkmark & \checkmark
    &0.382& 0.231 &0.153 &0.110 & 0.143 & 0.278 \\
    $ $ &\checkmark  &  &\checkmark  &\checkmark &0.379& 0.230 &0.154 &0.110 &0.144 &0.281\\
    \textbf{MIMIC-CXR}  &\checkmark  &\checkmark  &  &  & 0.337&   0.207&  0.139&  0.099 &  0.132& 0.273 \\ 
    &\checkmark  &\checkmark  & \checkmark  &  &0.361& 0.218 &0.146 &0.101 & 0.136 & 0.275 \\
    $ $ &\checkmark  &\checkmark  &  & \checkmark & 0.371&   0.224&  0.149&  0.108 &  0.141 & 0.278 \\
    \cline{2-11}
    $ $ &\checkmark  &\checkmark  &\checkmark  &\checkmark  & \textbf{0.393} & \textbf{0.243} & \textbf{0.159} &\textbf{0.113}& \textbf{0.160} &\textbf{0.285}   \\
    \bottomrule
  \end{tabular}
\end{table*}

\subsection{Comparison with state-of-the-art}
To demonstrate the effectiveness, we compare the performances of our model with a wide range of state-of-the-art models on the MIMIC-CXR and IU-Xray. \cref{tab:result_compare} shows the comparison results on both NLG and CE metrics. The models we compare to include Up-Down\cite{anderson2018bottom},  R2GenCMN\cite{chen2022cross}, PPKED\cite{liu2021exploring}, XPRONET\cite{wang2022cross}, \etal. 
As shown in \cref{tab:result_compare}, our KiUT outperforms state-of-the-art methods across all metrics on the MIMIC-CXR. Compared to the methods not specially designed for the medical domain\cite{anderson2018bottom,rennie2017self,chen2022cross}, our model shows a notable improvement. 
Furthermore, the result validates that our knowledge injection strategy, \ie, introducing a symptom graph in the final decoding step, 
performs better than those methods that are specially designed for radiology report generation\cite{liu2021exploring,you2021aligntransformer,liu2021contrastive,wang2022cross}. 
Moreover, the superior clinical efficacy scores that measure the accuracy of generated reports for clinical abnormalities demonstrate that our approach can produce higher-quality descriptions for clinical abnormalities.

Our model also significantly outperforms all the other methods in terms of the evaluation metrics on the IU-Xray datasets, except some benchmarks on the IU-Xray are slightly inferior to XPRONET\cite{wang2022cross}. This could be partly explained by XPRONET adopting cross-modal prototypes to record the information following \cite{chen2022cross}. The cross-modal module is easier to learn informative features in the small dataset IU-Xray, but fails to learn adequate information in the MIMIC-CXR whose size is almost 100 times larger than IU-Xray. As \cref{fig:compare} shows, the length of reports in MIMIC-CXR is longer than that in IU-Xray. This is further reflected in our proposed model has better generalizability.

\begin{table}[h!]
  \caption{Quantitative analysis for the U-connection structure. The experiments conducted on $\mathcal{M}^2 \text {Transformer}$, $\text{RSTNet}$, $\text{R2GenCMN (R2CMN)}$ and our $\text{KiUT}$. RSTNet and R2CMN adpot the last connection schema.}
  \label{tab:U-connection}
  \centering
  \resizebox{0.5\textwidth}{!}{
  \begin{tabular}{c|l|c|c|c|c}
    \toprule
   Dataset & Method  & Bleu1  &Bleu4 &Meteor & Rouge\_L \\
    \midrule
    &$\mathcal{M}^2 \text{Trans}$\cite{cornia2020meshed} & \textbf{0.808} &\textbf{0.391} &\textbf{0.292} &\textbf{0.586}\\
    &$\mathcal{M}^2 \text {Trans}^{{1-\text{to}-1}}$&0.803 &0.382 &0.289 &0.582 \\
    \textbf{COCO}&$\mathcal{M}^2 \text {Trans}^{{\text{U}}}$& \textbf{0.808}&  \textbf{0.391}  & 0.289  & 0.583 \\
    \cline{2-6}
    \textbf{test}\cite{lin2014microsoft}&$\text{RSTNet}$\cite{zhang2021rstnet} &0.811 &0.393 &\textbf{0.294} &0.588 \\
    &$\text{RSTNet}^{{1-\text{to}-1}}$& 0.809&  0.398  & 0.288  & 0.581 \\
    &$\text {RSTNet}^{{\text{U}}}$& \textbf{0.814} &\textbf{0.403} &0.293 &\textbf{0.591}\\
    \cline{1-6}
    &$\text{R2CMN}$\cite{chen2022cross}&  0.353&  0.103  &  \textbf{0.142} &\textbf{0.277} \\
    &$\text{R2CMN}^{{1-\text{to}-1}}$&  0.356&  0.101  &  0.141 & 0.273 \\
    \textbf{MIMIC}& $\text{R2CMN}^{{\text{U}}}$&  \textbf{0.363}&  \textbf{0.106}  & \textbf{0.142} & 0.276 \\
    \cline{2-6}
    \textbf{CXR}&$\text{KiUT}^{\text{last}}$&  0.372&  0.109  &  0.142 & 0.279 \\
    &$\text{KiUT}^{{1-\text{to}-1}}$& 0.380&  0.110  &  0.143 & 0.281 \\
    &$\text{KiUT}^{{\text{U}}}$
     & \textbf{0.393}  &\textbf{0.113}& \textbf{0.160} &\textbf{0.285}  \\
    \bottomrule
  \end{tabular}
  }
\end{table}

\subsection{Ablation Studies}
To fully investigate the contribution of our proposed Region Relationship Encoder(RR-Encoder), Injected Knowledge Distiller(IK-Distiller), and the U-connection schema. We conduct a quantitative analysis to compare each component. The main results are shown in \cref{tab:ablation} and \cref{tab:U-connection}.

\textbf{Effect of RR-Encoder} We start from a basic encoder without the extrinsic relationship (ER) and the intrinsic relationship (IR). Then we add ER and IR, respectively. Comparing the results in \cref{tab:ablation}, both ER and IR can boost the performance for the base transformer encoder, \eg, 0.373 $\rightarrow$ 0.379 and 0.373 $\rightarrow$ 0.382 in BLEU-1 score respectively. As the results show, IR brings a more considerable improvement than ER (0.382 BLEU-1 vs. 0.379 BLEU-1), which indicates that exploiting the intrinsic relationship is effective when encoding radiology image regions. The performance gain of RR-Encoder comes from the similar anatomical structure of human organs, in which the extrinsic and intrinsic relationship is beneficial for extracting the inherent information of chest X-rays images.

\begin{figure*}[t]
\centering
\scalebox{0.97}{
\includegraphics[width=\linewidth]{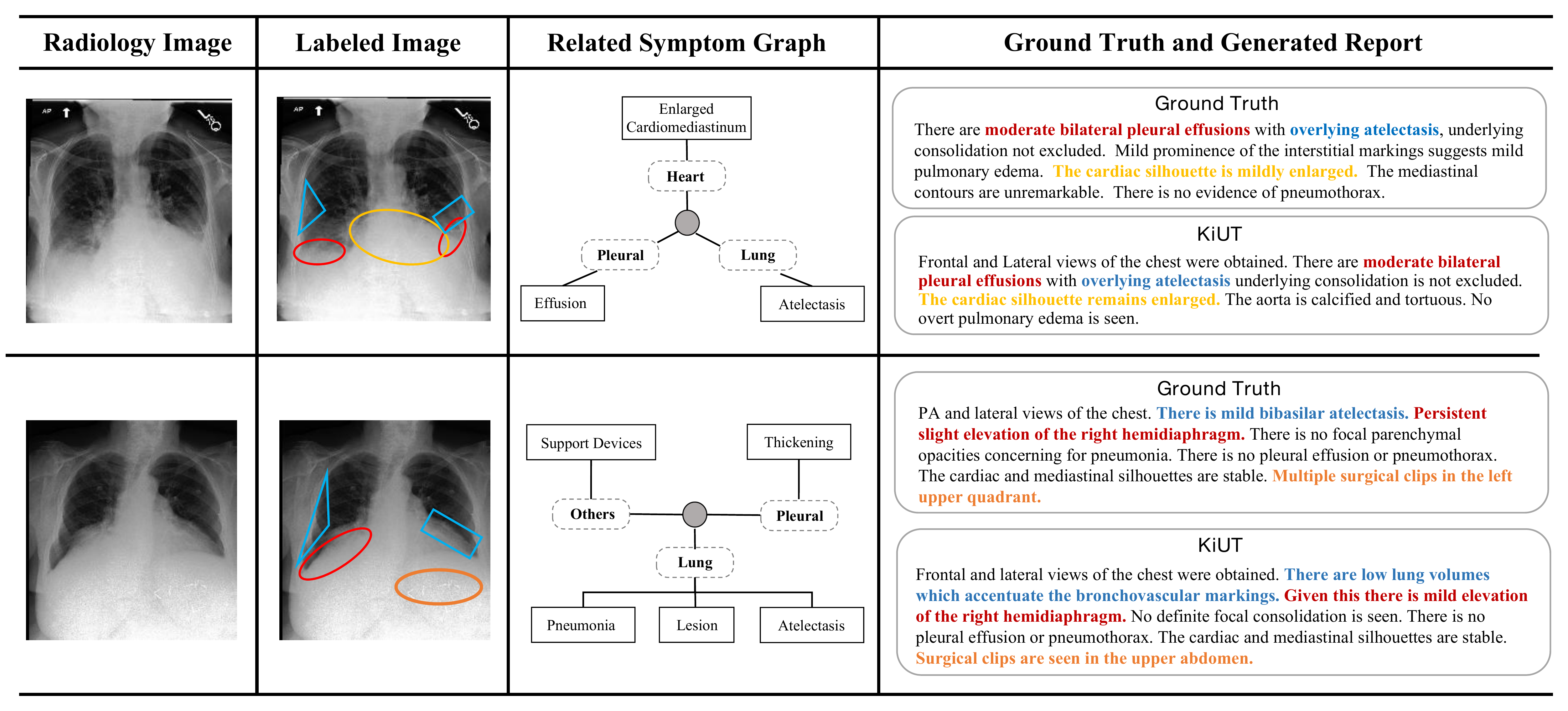}}
\caption{The visualization of the \emph{KiUT}. The colored bounding regions in labeled and colored text in reports represent the abnormalities. The symptom graph represents a symptom subgraph inferred from the input image and extracted in our graph. Different colors indicate different symptoms. Different bounding shapes of one symptom mean different medical manifestations in clinical diagnosis.}
\label{fig:vis}
\vspace{-1em}
\end{figure*}

\textbf{Effect of IK-Distiller} We evaluate the impact of the proposed contextual knowledge and clinical knowledge in IK-Distiller. Specifically, the distiller without contextual feature and clinical feature is a transformer-based decoder layer. \cref{tab:ablation} shows these two additional knowledge feature significantly boost the performance as a dramatic reduction can be seen when both of them are removed (from 0.393 to 0.337). 
In detail, the contextual feature is essential in generating long free text, such as radiology reports, since there are a certain number of tokens in a long text that is not related to the input images and should be inferred from the contexts. For the clinical feature, it helps the model to obtain a radiologist-like ability with external knowledge in generating the description of the abnormalities. With clinical features, the decoding process can be regarded as a report writing process of a radiologist, who examines the image first and employs the related clinical knowledge to complete a report.  Overall, since contextual features and clinical features benefit the performance from different perspectives, adopting IK-Distiller can lead to generating more accurate reports with abnormalities descriptions.

\textbf{Analysis of Connection} We evaluate the role of the U-connection between the encoder and decoder layers. In the previous work, the transformer-based model has been successfully applied to the captioning tasks with the original connection structure for uni-modal scenarios like machine translation. We speculate that the generating task in cross-modal scenarios requires more specific architectures. Thus we propose U-connection and compare variations of the original Transformer connection structure with it. As shown in \cref{tab:U-connection}, we introduce different connection methods in natural image caption models and radiology report generation models. It can be seen that the 1-to-1 connection schema where the $i$-th decoder layer is only connected to the corresponding $i$-th encoder layer can bring an improvement with respect to only applying the output of the last encoder layer. Furthermore, our proposed U-connection can boost the performance of R2GenCMN (from 0.353 to 0.363 in the BLEU-1 score). More encouragingly, the U-connection schema can achieve comparable performances as the meshed connection\cite{cornia2020meshed} with a simple structure and fewer parameters.  
Thus, we confirm that exploiting the interaction between the encoder and decoder is beneficial, and the U-connection is more conformable in the cross-modal scenario than the other connection schema.

\subsection{Qualitative Analysis}

To better understand the effectiveness of our model, qualitative examples are given in \cref{fig:vis}. Intuitively, the reports generated by \emph{KiUT} are accurate and robust, which show significant alignment with ground truth reports. As the figure shows, our model can learn from radiology image features and extract the relevant symptoms from the symptom graph (\eg, ``enlarged cardiomediastinum'', ``pleural effusions'' and ``atelectasis''). 
It is worth noting that our model can find and describe some subtle observations, such as \textit{``support devices''} and \textit{``slight elevation of the right hemidiaphragm''}. This could be attributed to the well-designed U-connection for multi-level interaction and the injected knowledge. 
On the other hand, these examples indicate the ability of our model to accurately discriminate and describe the different clinical manifestations of the same symptom (the blue labels). Owing to the employment of injected knowledge distiller, our model also has the certain ability to reason and write. As the example shown in \cref{fig:introduction} (\textit{GT: ``The cardiac silhouette also appears enlarged, but it is difficult to completely assess the left border given the large pleural effusion.''} and \textit{Gen: ``The cardiac silhouette is difficult to assess due to the left base opacity.''}), the sentence should be generated according to the context and the correspondence of different symptoms. More visualizations are included in the supplementary material.
\section{Conclusion}
In this paper, we present \emph{KiUT}, a novel framework for radiology report generation that focuses on extracting and distilling multi-level information and multiple injected knowledge. Our model encodes images with the extrinsic and intrinsic relationships among image regions and decodes words through an injected knowledge distiller. We propose a novel U-connection schema to exploit the interaction between the encoder and decoder, which is unprecedented for other architectures in such a cross-modal scenario. Experimental results on the MIMIC-CXR and IU-Xray datasets demonstrate that our approach achieves a state-of-the-art performance and outperforms recent research from the external knowledge view. Ablation studies also prove the effectiveness of the proposed parts. We leave sophisticated knowledge constructing and structured report template filling solutions as future work.

{\small
\bibliographystyle{ieee_fullname}
\bibliography{egbib}
}

\end{document}


\title{Supplementary Materials}  

\maketitle
\thispagestyle{empty}
\appendix


\section{Details of Visualization}
\begin{figure*}[htb]
\centering
\scalebox{1}{
\includegraphics[width=\linewidth]{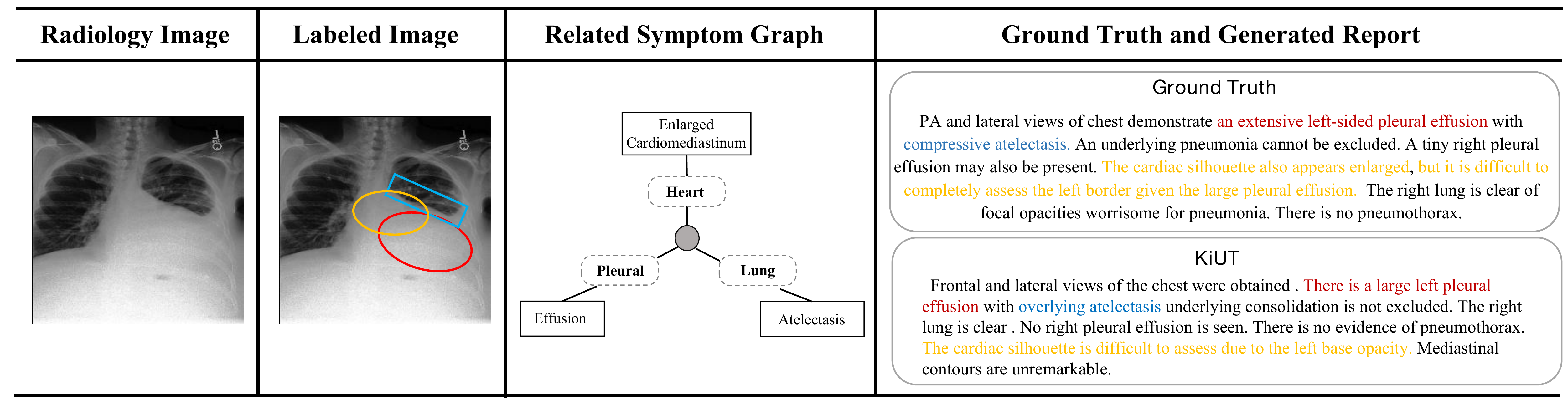}}
\caption{The visualization example introduced in the \textbf{Figure 1} and \textbf{Section 4.6} of \emph{KiUT} of the paper.}
\label{fig:vis_intro}
\end{figure*}

\begin{figure*}[htb]
\centering
\scalebox{1}{
\includegraphics[width=\linewidth]{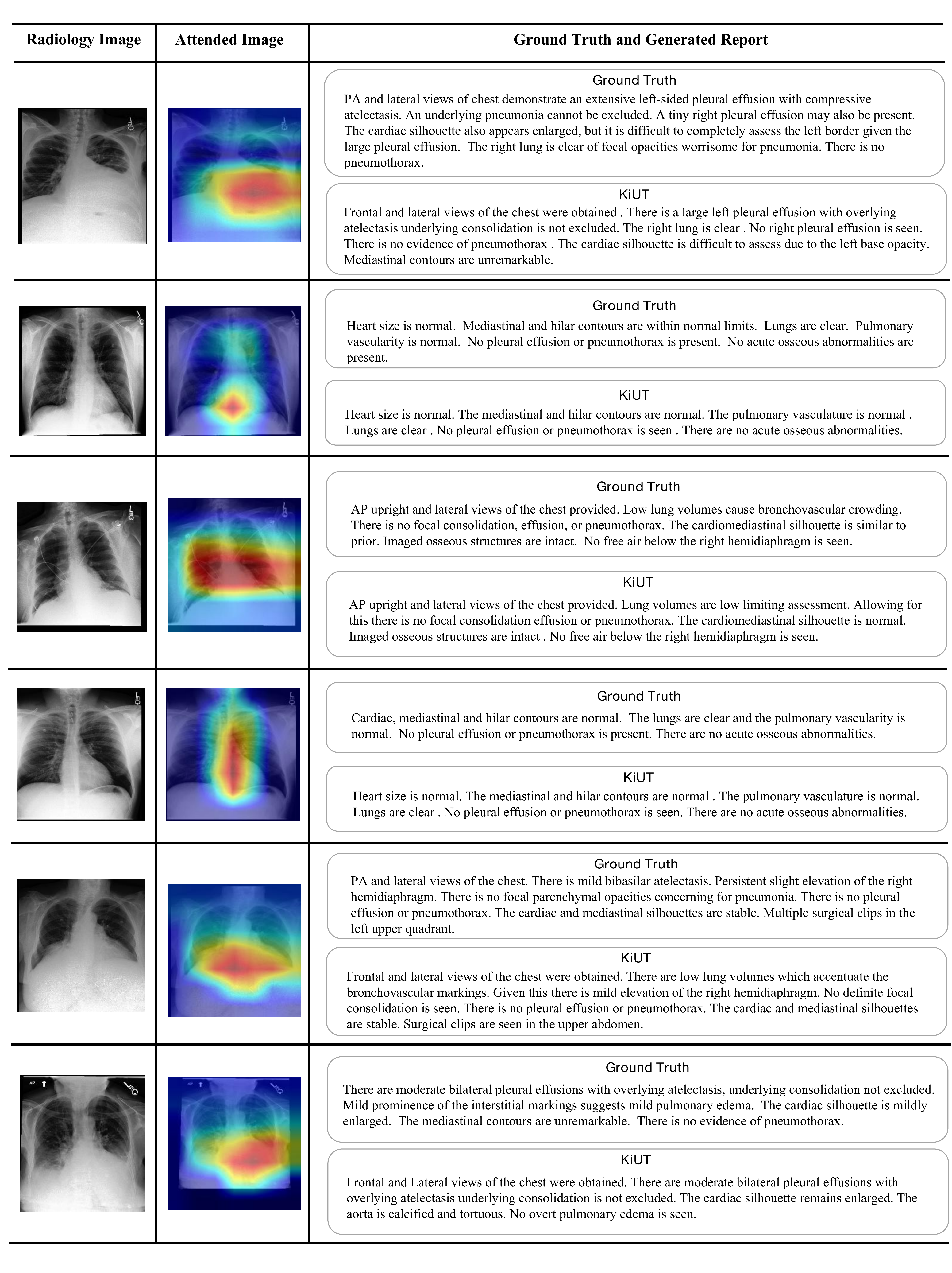}}
\caption{The visualization examples of our \emph{KiUT}.}
\label{fig:morevis}
\end{figure*}

\cref{fig:vis_intro} shows the detailed generated report of \emph{KiUT} which is introduced in \textbf{Section 4.6}. And  \cref{fig:morevis} gives more examples of our \emph{KiUT}.

\section{Experiments}
Intuitively, the different number of transformer layers with varying connection schemas can exhibit different results. It is easy to deduce that the intricacy of connection schema does not affect the performance of models with few layers, \eg, one layer. However, a properly designed connection schema between the encoder and decoder with multiple layers could achieve better performances without any bells and whistles. To investigate the impact of the number of encoding and decoding layers in our U-transformer, we report the performances of models with the different numbers of layers,  as it can be seen in \cref{tab:layers}.

\begin{figure}[htb]
\centering
\scalebox{0.8}{
\includegraphics[width=\linewidth]{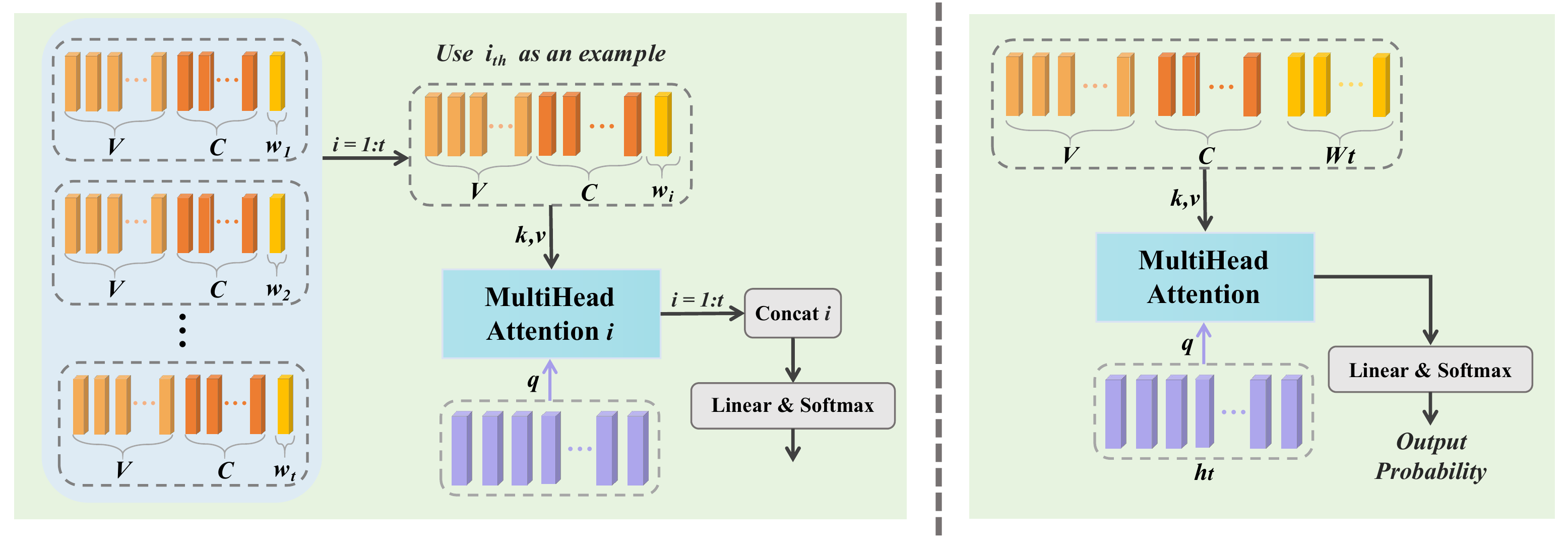}}
\caption{Different ways to distill knowledge}
\label{fig:diff}
\end{figure}

\begin{table}[h!]
  \caption{Quantitative analysis for the number of transformer layers with different schema. The experiments are conducted on the IU-Xray. The symbol $∗$ denotes our replicated results with the official codes. The best results are underlined}
  \label{tab:layers}
  \centering
  \begin{tabular}{c|l|c|c|c|c|c|c}
  \midrule
   Dataset & Method  & Bleu1 &Bleu2 &Bleu3&Bleu4 &Meteor & Rouge\_L \\
    \cline{1-8}
    & $\text{R2GenCMN}^{*}$(3 layers) & 0.450&  0.292  &  0.213 &0.166 &0.184 &0.350 \\
    &$\text{R2GenCMN}^{*}$(6 layers) & 0.455&  0.285  &  0.202 &0.147 &0.189 &0.359 \\
    &$\text{R2GenCMN}^{{\text{U}}}$(3 layers) & \underline{0.488}&  \underline{0.312}  &  0.222 &0.167 &\underline{0.197} &\underline{0.364} \\
    \textbf{IU-}&$\text{R2GenCMN}^{{\text{U}}}$(6 layers) & 0.475&  0.306  &  \underline{0.224} &\underline{0.174} &0.189 &0.363 \\
    \cline{2-8}
    \textbf{Xray}& $\text{KiUT}^{{\text{last}}}$(3 layers) & 0.483&  0.323  &  0.226 &0.170 &0.225 &0.382 \\
    &$\text{KiUT}^{{\text{last}}}$(6 layers) & 0.489&  0.324  & 0.225& 0.166 &0.221 &0.379\\
    &$\text{KiUT}^{{\text{U}}}$(3 layers) & \underline{0.525}&  \underline{0.360}  &  \underline{0.251} &\underline{0.185} &\underline{0.242} &\underline{0.409} \\
    &$\text{KiUT}^{{\text{U}}}$(6 layers) & 0.519&  0.341  &  0.229 &0.165 &0.229 &0.400 \\
    \bottomrule
  \end{tabular}
\end{table}
We further compare the results of different ways of distilling injected knowledge. As \cref{tab:ddd} shows, the way adopted in \textit{KiUT}(the left figure in \cref{fig:diff}) can surpass the right one in \cref{fig:diff}.

\begin{table}[h!]
  \caption{Quantitative analysis of different ways for distilling injected knowledge.}
  \label{tab:ddd}
  \centering
  \begin{tabular}{c|l|c|c|c|c|c|c}
  \midrule
   Dataset & Method  & Bleu1 &Bleu2 &Bleu3&Bleu4 &Meteor & Rouge\_L \\
    \cline{1-8}
    \textbf{MIMIC}&$\text{KiUT}^{\text{right}}$& 0.384&  0.235  &  0.150 &0.106 &0.155 &0.278 \\
    \cline{2-8}
    \textbf{CXR}&$\text{KiUT}^{\text{left}}$
     & \textbf{0.393}  & \textbf{0.243} & \textbf{0.159} &\textbf{0.113}& \textbf{0.160} &\textbf{0.285}  \\
    \bottomrule
  \end{tabular}
 
\end{table}

As shown in \cref{tab:more_result_compare}, we also compare our approach with more state-of-the-art radiology report generation models proposed in the early exploration phase of the task, \ie, HRNN\cite{krause2017hierarchical}, HRGR-Agent\cite{li2018hybrid}, CoAtt\cite{jing2017automatic}, CMAS-RL\cite{jing2020show} in IU-Xray and CNN-RNN\cite{vinyals2015show},
AdaAtt\cite{lu2017knowing}, Att2in\cite{rennie2017self} and Up-Down\cite{anderson2018bottom}.

\begin{table*}[htb]
  \caption{More detailed performance comparisons of the proposed KiUT with more existing methods on the test sets of MIMIC-CXR and IU-Xray datasets with respect to NLG and CE metrics. The best values are highlighted in bold.}
  \label{tab:more_result_compare}
  \centering
  \begin{tabular}{c|l|c|c|c|c|c|c|c|c|c}
    \hline
    \multicolumn{1}{c|}{\multirow{2}{*}{Dataset}} &
    \multicolumn{1}{c|}{\multirow{2}{*}{Method}} &
    \multicolumn{6}{|c}{NLG Metric} & 
    \multicolumn{3}{|c}{CE Metric} 
    \\\cline {3-11} ~ & ~ & Bleu1 & Bleu2 &Bleu3 &Bleu4 &Meteor & Rouge\_L  & Precision & Recall & F1 \\ \hline \hline
    $ $ &HRNN\cite{krause2017hierarchical} &0.439 &0.281 &0.190 &0.133 &$-$ &0.342 & $-$& $-$ & $-$\\
    $ $ &HRGR-Agent\cite{li2018hybrid} &0.438 &0.298 &0.208 &0.151 &$-$ &0.322 & $-$& $-$ & $-$\\
    $ $ &CoAtt\cite{jing2017automatic} &0.455 &0.288 &0.205 &0.154 &$-$ &0.369 & $-$& $-$ & $-$\\
    $ $ &CMAS-RL\cite{jing2020show} &0.464 &0.301 &0.210 &0.154 &$-$ &0.362 & $-$& $-$ & $-$\\
    \textbf{IU-} &SentSAT+KG\cite{zhang2020radiology} &0.441 &0.291 &0.203 &0.147 &$-$ &0.367 & $-$& $-$ & $-$\\
    \textbf{Xray}  & R2GenCMN\cite{chen2022cross}&0.470&0.304&0.219&0.165&0.187&0.371& $-$ & $-$ & $-$\\
    $ $ & PPKED\cite{liu2021exploring}&0.483&0.315&0.224&0.168&0.190& 0.376 & $-$ & $-$ & $-$\\
    $ $ & AlignTrans\cite{you2021aligntransformer}& 0.484&  0.313& 0.225& 0.173& 0.204& 0.379  & $-$ & $-$ & $-$\\
  $ $ &Contrastive\cite{liu2021contrastive} &0.492 &0.314 &0.222 &0.169 &0.193 &0.381 & $-$& $-$ & $-$\\
    $ $ & XPRONET\cite{wang2022cross} & 0.525 & 0.357 & \textbf{0.262} &\textbf{0.199}& 0.220 & \textbf{0.411}  &$-$ & $-$ & $-$\\
    \cline {2-11}
  $ $ & Ours& \textbf{0.525} & \textbf{0.360} & 0.251 & 0.185& \textbf{0.242} &0.409 & $-$ & $-$ & $-$\\
    \hline
      $ $ & CNN-RNN\cite{vinyals2015show}&0.299&	0.184&	0.121&	0.084&	0.124&	0.263  &  0.249 & 0.203& 0.204\\
    $ $ & AdaAtt\cite{lu2017knowing}& 0.299&   0.185&  0.124&  0.088&  0.118 & 0.266  & 0.268 & 0.186 & 0.181\\
    $ $& Up-Down\cite{anderson2018bottom}&  0.317&   0.195&  0.130&  0.092&  0.128 & 0.267 &  0.320 &0.231 &0.238\\
    \textbf{MIMIC} & Att2in\cite{rennie2017self}&  0.325&   0.203&  0.136&  0.096 &  0.134 & 0.276  &  0.322 & 0.239 & 0.249\\
    \textbf{CXR} & R2GenCMN\cite{chen2022cross}&  0.353&    0.218&   0.145&  0.103&   0.142& 0.277  & 0.333 &0.273 &0.276\\
    $ $ & PPKED\cite{liu2021exploring}& 0.360&    0.224&   0.149&  0.106&  0.149& 0.284 & $-$ & $-$ & $-$\\
    $ $ & Contrastive\cite{liu2021contrastive}& 0.350 &0.219 &0.152  &0.109 &0.151 &0.283 &0.352 &0.298 &0.303\\
    $ $ & AlignTrans\cite{you2021aligntransformer}& 0.378&  0.235& 0.156& 0.112& 0.158& 0.283  & $-$ & $-$ & $-$\\
    $ $ & XPRONET\cite{wang2022cross} & 0.344 & 0.215 & 0.146 & 0.105& 0.138 & 0.279  & $-$ & $-$ & $-$\\
    \cline {2-11}
    $ $ & Ours& \textbf{0.393} & \textbf{0.243} & \textbf{0.159} &\textbf{0.113}& \textbf{0.160} &\textbf{0.285} & \textbf{0.371} & \textbf{0.318} & \textbf{0.321}\\
    \bottomrule
  \end{tabular}
\end{table*}

\newpage
\section{Details of \textit{KiUT}}
\subsection{The RR-Encoder}
We input the $X = \{x_i\}_{i=1}^{S}, x_i \in \mathbb{R}^d$ extracted from backbone into our Region Relationship Encoder. The patches are then passed through the encoder consisting of a sequence of $N$ region relationship augmented self-attention$(\mathcal{RRSA})$  based transformer layers. Each layer $\ell$ comprises of $(\mathcal{RRSA})$, layer normalization($\operatorname{LN}$), and $\operatorname{MLP}$ blocks as follows:
\begin{gather}
X^{'\ell} =\operatorname{MRRSA}\left(\operatorname{LN}\left(X^{\ell}\right)\right)+X^{\ell} \\
X^{\ell+1} =\operatorname{MLP}\left(\operatorname{LN}\left(X^{'\ell}\right)\right)+X^{'\ell}
\end{gather}
The MLP consists of two linear projections separated by a non-linearity activation function $\operatorname{GELU}$, and the dimension of $x_i \in X$, $d$, remains fixed throughout all layers.

\subsection{The Decoder}
After receiving the visual features $\lbrace{\hat{x}_1,\dots,\hat{x}_N}\rbrace$ from the encoder, the decoder generates the hidden state $h_t$ to predict the current word $y_{t+1}$ through the word sequence features $\hat{w}_{t}$ in the decoding step t+1. Given the generated words $y_{1:t} = \lbrace{y_1, \dots, y_t}\rbrace$, the word sequence features are obtained by words embeddings $w_t$ combining with its position encodings $e_t$.
The $i$-th decoder layer based on the Transformer layer can be formulated as:
\begin{gather}
\operatorname{PosEncoding} = \begin{cases}
\operatorname{PE}{(\text {pos}, 2k)} =\sin \left(pos/ 10000^\frac{2k}{d_n}\right) \\
\operatorname{PE}{(\text {pos}, 2k+1)} =\cos \left(pos / 10000^\frac{2k}{d_n}\right)
\end{cases}\\
e_{t} = \operatorname{PosEncoding}(y_{1:t})\\
\hat{w}_{t} = w_{t}+e_{t} \\
\hat{h}^i_t =\begin{cases} \operatorname{AddNorm}(\operatorname{MHA}(\hat{w}_{t},\hat{w}_{t})) & i=1 \\
\operatorname{AddNorm}(\operatorname{MHA}(h^{i-1}_t,h^{i-1}_t)) & i>1 \\
\end{cases}
\\ \tilde{h}^i_t = \operatorname{AddNorm}(\operatorname{MHA}(\hat{h}^i_t,\hat{x}_i)
\\h^i_t=\operatorname{AddNorm}(\mathcal{FFN}(\tilde{h}^i_t))
\\\operatorname{MHA}(X, Y)=\left[\operatorname{Att}_{1}(X, Y) , \ldots , \operatorname{Att}_{n}(X, Y)\right] \mathrm{W}^{\mathrm{O}}
\\\operatorname{Att}_{i}(X, Y)=\operatorname{Softmax}\left(\frac{X \mathrm{~W}_{i}^{\mathrm{Q}}\left(Y \mathrm{~W}_{i}^{\mathrm{K}}\right)^{T}}{\sqrt{d_{n}}}\right) Y \mathrm{~W}_{i}^{\mathrm{V}}
\\\mathcal{FFN}(x)=\max \left(0, x \mathrm{~W}_{0}+\mathrm{b}_{0}\right) \mathrm{W}_{1}+\mathrm{b}_{1}\end{gather}
where ${pos}$ is the position and $k$ is the dimension, the $\mathrm{W}_{i}^{\mathrm{Q}}, \mathrm{W}_{i}^{\mathrm{K}}, \mathrm{W}_{i}^{\mathrm{V}} \in$ $\mathbb{R}^{d \times d_{n}}$ and $\mathrm{W}^{\mathrm{O}} \in \mathbb{R}^{d \times d}$ are learnable parameters, $d_{n}=d / n$, $[\cdot , \cdot]$ stands for concatenation operation,  $\mathrm{W}_{0} \in \mathbb{R}^{d \times 4 d}$ and $\mathrm{W}_{\mathrm{1}} \in \mathbb{R}^{4 d \times d}$ denote learnable matrices for linear transformation, and $b_{0}$ and $b_{1}$ are the bias terms.

{\small
\bibliographystyle{ieee_fullname}
\bibliography{egbib}
}